\pdfoutput=1

\documentclass[11pt]{article}

\usepackage[]{emnlp2021}

\usepackage{times}
\usepackage{latexsym}

\usepackage{amsmath}
\usepackage{amsthm}
\usepackage{booktabs}
\usepackage{graphicx}
\usepackage{multirow}
\usepackage{algorithm}
\usepackage{algorithmic}
\usepackage{amsfonts,amssymb}
\usepackage{bm}
\usepackage{enumitem}
\usepackage[T1]{fontenc}

\usepackage[utf8]{inputenc}

\usepackage{microtype}

\title{First Target and Opinion then Polarity: A Two-stage Correlation Enhanced Network for Aspect Sentiment Triplet Extraction}

\author{\textbf{Lianzhe Huang}$^{1}$\thanks{~~Equal contribution.}, \textbf{Peiyi Wang}$^{1*}$, \textbf{Sujian Li}$^{1}$, \textbf{Tianyu Liu}$^{1}$ \\ \textbf{Xiaodong Zhang}$^{2}$, \textbf{Zhicong Cheng}$^{2}$, \textbf{Dawei Yin}$^{2}$, \textbf{Houfeng Wang}$^{1}$\\ 
$^1$MOE Key Lab of Computational Linguistics, Peking University, Beijing, 100871, China \\
$^2$Baidu Inc.\\
\texttt{\{hlz, lisujian, tianyu0421, wanghf\}@pku.edu.cn} \\ \texttt{wangpeiyi@stu.pku.edu.cn} \\
\texttt{\{zhangxiaodong11, chengzhicong01\}@baidu.com},  
\texttt{yindawei@acm.org}
}

\begin{document}
\maketitle

\begin{abstract}
Aspect Sentiment Triplet Extraction (ASTE) aims to extract triplets from a sentence, including target entities, associated sentiment polarities, and opinion spans which rationalize the polarities.
Existing methods are short on building correlation between target-opinion pairs, and neglect the mutual interference among different sentiment triplets.
To address these issues, we utilize a two-stage framework to enhance the correlation between targets and opinions: at stage one, we extract targets and opinions through sequence tagging; then we append a group of artificial tags named Perceivable Pair, which indicate the span of a specific target-opinion tuple, to the input sentence to obtain closer correlated target-opinion pair representation. Meanwhile, we reduce the negative interference between triplets by restricting tokens' attention field. Finally, the polarity is identified according to the representation of the Perceivable Pair. 
We conduct experiments on four datasets, and the experimental results show the effectiveness of our model. 
\end{abstract}
\section{Introduction}\label{chap: intro}

\begin{figure}[t]
    \centering
    \includegraphics[width=0.5\textwidth]{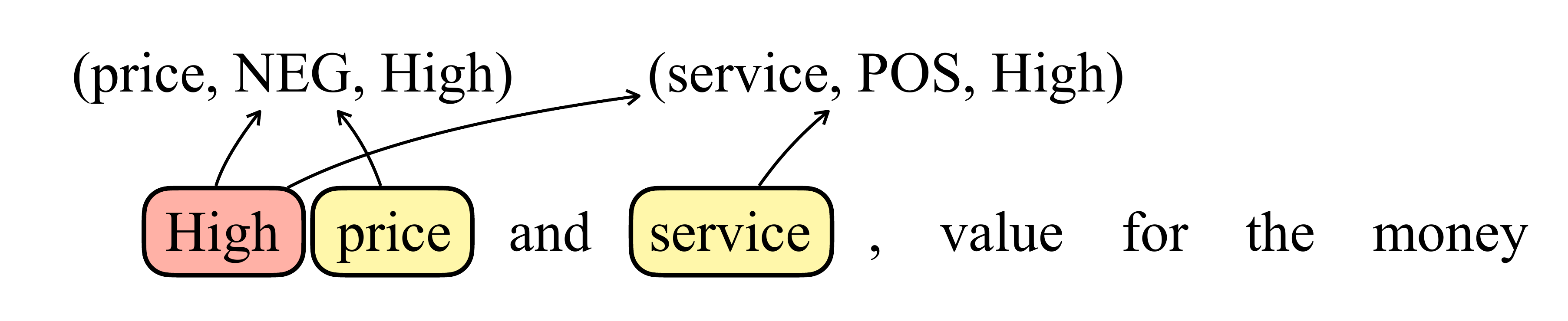}
    \caption{An example of ASTE. The target words are colored yellow, and the opinion words are colored red.}
    \label{fig:data_fig}
\end{figure}

\textbf{A}spect-\textbf{B}ased \textbf{S}entiment \textbf{A}nalysis (\textbf{ABSA}) \cite{liu2012sentiment,ma2017interactive,zhao2019modeling} task aims at identifying the sentiment 
associated with a specified target in a sentence.
It constitutes a corner stone for sentiment analysis \cite{feldman2013techniques,zhang2018deep} applied in different scenarios such as social media \cite{agarwal2011sentiment}, e-commerce \cite{fang2015sentiment}, and the press industry \cite{godbole2007large}. 
In this paper, we focus on the \textbf{A}spect \textbf{S}entiment \textbf{T}riplet \textbf{E}xtraction (\textbf{ASTE}) task. ASTE is expanded from ABSA however has two major differences: $i)$ ASTE does not specify target entities in the sentence, instead practitioners are required to exhaustively extract all entities along with related sentiment. $ii)$ For all extracted sentiment, the corresponding rationale, which we call ``opinion'', should also be included in the final output. In Figure \ref{fig:data_fig}, the opinion word ``\textit{High}'' indicates a \textit{negative} attitude towards the target ``\textit{price}'' in the triplet {(price, NEG, High)}.

The prior work can be roughly divided into two categories, namely tagging-based methods and matching-based methods.
To assign proper sentiment polarities to the corresponding target-opinion pairs, a principled tagging-based method \cite{xu2020position} allocates a composite tag (e.g. a tuple) to each word in the sentence, which indicates the positions of target and opinion words and related sentiment. 
But such tagging methods fail to handle the one-opinion many-targets situations as illustrated in Figure \ref{fig:data_fig}. 
Different from the tagging-based methods, 
the matching-based methods adopt a ``first extract and then match'' pipeline.
Different extraction methods correspond to specific model configurations on subtask combination, e.g.
\citet{peng2020knowing} extracts target-polarity tuples and opinion words while \citet{zhang2020multi} obtain target words, opinion words and sentiment polarities separately via a multi-task learning framework.
Both \citet{peng2020knowing} and \citet{zhang2020multi} perform pairwise matching after extraction.

The potential risks in prior work include $i)$ the defective subtask combination. According to human perception, people naturally judge the sentiment based on target-opinion pairs, e.g. negative for ``high price'' and positive for ``high service''. Comparing with independently retrieving target, opinion and sentiment opponents \cite{zhang2020multi} or matching target-polarity tuples with opinion words \cite{peng2020knowing}, obtaining target-opinion tuples at the first stage and then assigning corresponding sentiment polarity would be a better subtask combination.
$ii)$ the inefficiency in handling ``many targets to one opinion'' and ``many opinions to one target'' situations. 
Taking the opinion word ``high'' in Figure \ref{fig:data_fig} as an example, in phrase ``high price'' it means ``expensive'', when paired with ``service'', ``high'' refers to ``good quality''.
However in prior work \cite{peng2020knowing,zhang2020multi}, they first obtain the representation of word ``high'' through 
sentence encoders. 
Then while matching targets ``service'' or ``price'', they use the \textit{same} representation for the opinion word ``high'' regardless of the ambiguity in different target-opinion word collocations.

To this end, we use a two-stage framework that first extracts target and opinion words in the sentence and then judges the sentiment polarity with target-opinion-tuple-aware representation. 
The subtask setting adheres to human cognition, i.e. making the judgement (sentiment polarity) according to the supporting evidence (target-opinion pairs). 
To mitigate the ambiguity in different target-opinion word collocation, we enumerate all possible target-opinion pairs and append them to the input sentence.
In the ``many targets to one opinion'' and ``many opinions to one target'' situations, while matching different targets and opinions, we would choose the corresponding pair representation in the input sequence.

Specifically in the first stage, we exploit a BERT-based \cite{devlin2018bert}
tagging model to extract the target words and opinion words from input sentences like previous work \cite{peng2020knowing,xu2020position}. 
But we do not perform any other additional tasks such as sentiment classification like them at this stage. 
In the second stage, 
inspired by the mark tokens in the relation classification task~\cite{zhang2019ernie,soares2019matching,zhong2020frustratingly}, 
we use a group of artificial tags to form a specific \textit{Perceivable Pair} for each target-opinion pair and append them to the input sentence. 
Follow the settings of \cite{zhong2020frustratingly}, 
the \textit{perceivable pairs} share the same position embedding with the related words in the input sentence to explicitly point out the positions of target and opinion spans. 
With BERT encoder, when determining the sentiment polarity for a potential target-opinion pair, we retrieve the corresponding perceivable pair representation from the input sequence for 4-way sentiment classification. Apart from \textit{positive, neutral, negative} labels, we assign ``\textit{N/A}'' labels to the target-opinion pairs which can not constitute sentiment triples.
As there may be many target-opinion pairs in the sequence. To avoid negative interference between them, we adopt the attention constraint from \cite{zhong2020frustratingly}.
So when encoding a certain target-opinion pair in the input sequence, the model can only leverage the sentence representation and does not have the access to other target-opinion pairs to avoid negative interference. 
We summarize our contributions as follows:
\begin{itemize}
    \item We use a two-stage framework in ASTE which extracts target-opinion pairs and then judges their sentiment polarity. The subtask combination achieves good performance and comports with human cognition.
    \item We utilize sequential input representation with perceivable pairs and restricted attention to enhance the correlation between target-opinion pairs, which resolves ambiguity in word collocation.
    \item We conducted experiments on four datasets, the experimental results show the effectiveness of our model. In addition, empirically our model performs better in complex situations like ``many-to-one'' relations and multiple triplets in one sentence.
\end{itemize}
\begin{figure*}[t]
    \centering
    \includegraphics[width=0.9\textwidth]{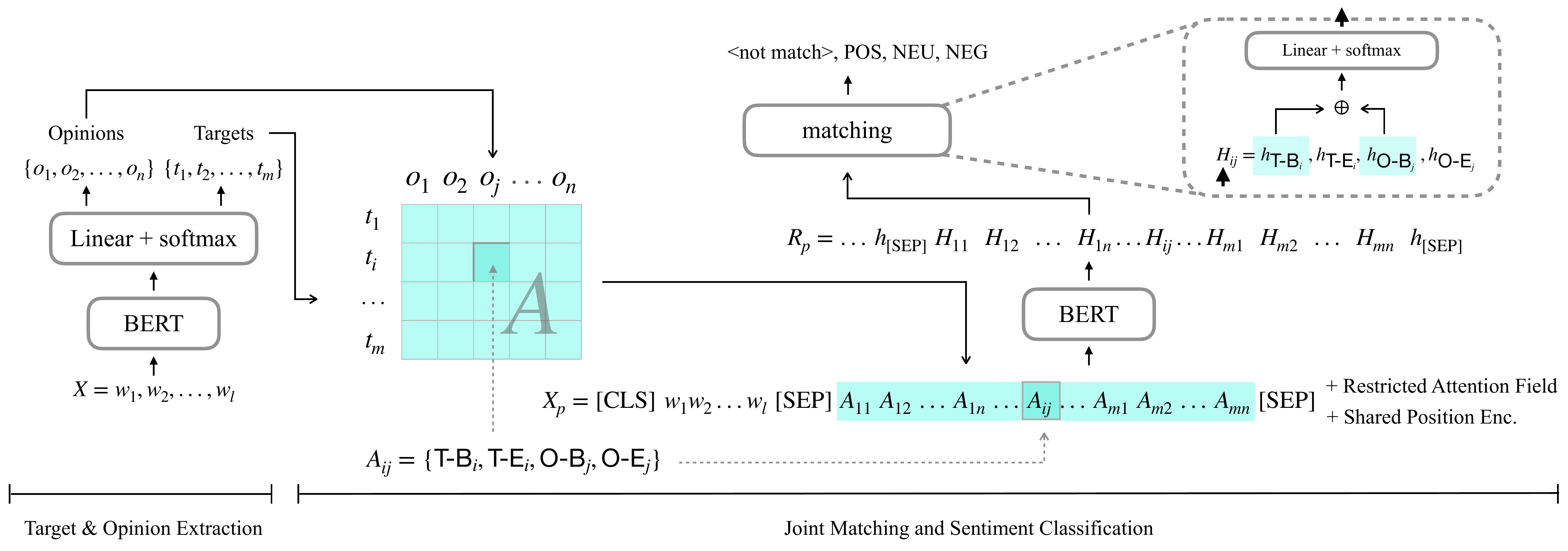}
    \caption{Overview of our approach. In the target and opinion extraction stage, we tag all target and opinion words in the sentence through a sequence labeling model. Then we assign a group of artificial tags to form a specific perceivable pair $A_{ij}$ for each possible target $i$ and opinion $j$ pairs. All the perceivable pairs form the $A$-matrix are appended to the end of the original input in order. This sequence will be sent into another BERT with restricted attention field and shared position embedding. Then put the BERT's output corresponding $A$-matrix into the matching component to get the final result. }
    \label{fig:main}
\end{figure*}

\section{Related Work}
In this section, we will briefly review works on sentiment analysis and triplet extraction.

Aspect-level Sentiment Analysis (ABSA) proposed by \cite{sentiment2014} has recently been receiving attention from the research community. 
In ABSA, many different kinds of tasks have been developed.
One is defined as analyzing a specified aspect's sentiment polarity in a sentence \cite{dong2014adaptive,ma2017interactive,tang2020dependency}. 
Other two kinds of sentiment analysis do not specify the aspects in advance, 
which has a strong similarity with Aspect Sentiment Triplet Extraction (ASTE). Aspect-sentiment pair extraction \cite{li2019unified}, which extracts the aspect word and its sentiment polarity from the sentence. And aspect-opinion co-extraction \cite{li2018aspect} tends to extract aspects and opinions simultaneously. The above two tasks are both subsets of the ASTE task. The ASTE task can analyze the sentiment polarity of the target, and give opinion words as the classification basis, which is more practical in actual scenarios.

In addition to the ASTE, another influential triplet extraction task in the NLP community is Joint Entity and Relation Extraction (JERE). The goal of JERE is to extract entities from sentences and give the relationships between the extracted entities. In terms of model granularity when matching, the existing work on JERE can be divided into token level models \cite{zhang2017end,zheng2017joint} and span level models  \cite{wadden2019entity,zhong2020frustratingly}. 
The Joint Matching and Sentiment Classification part of our framework is inspired by those span level models, emphasizing that spans need to be explicitly recognized and perceived by the model. 
The most similar work with us in JERE is ~\citet{zhong2020frustratingly}, we list the main differences between our framework and theirs from the perspective of the model structure below: {$i)$} the entities of the their JERE task may overlap, they use a span-based method for entity extraction, which is relatively inefficient because it requires searching all possible spans in the sentence. Since there is no overlap between target and opinion in our task,  we adopt token classification method to obtain the range of target and opinion. In the ASTE task, this will be more efficient than their method.
{$ii)$} Our marker tokens will form the Perceivable Pairs based on the pairing of target and opinion, rather than the simple pairwise combinations.
{$iii)$} We use different segment id to distinguish the original input and marker combinations (Perceivable Pair) part in matching stage. 
{$iv)$} They put all possible marker combinations in the sentence at one time only at inference time but not at training time due to model performance. In our framework, we put all possible the marker combinations (Perceivable Pair) in the sentence at the same time during training and does not affect the performance of the model, so all sentence is only calculate once during training and inference.

\begin{table*}[t]
\centering
\scalebox{0.7}{
\begin{tabular}{lrrrr|rrrr|rrrr|rrrr}
\toprule 
\multirow{2}{*}{\textbf{Split}} & \multicolumn{4}{c}{\texttt{14Rest}} & \multicolumn{4}{c}{\texttt{14Lap}} & \multicolumn{4}{c}{\texttt{15Rest}} & \multicolumn{4}{c}{\texttt{16Rest}} \\  
& Sent. & Pos. & Neu. & Neg. &  Sent. & Pos. & Neu. & Neg. & Sent. & Pos. & Neu. & Neg. & Sent. & Pos. & Neu. & Neg.  \\ \midrule
\textbf{Train} & 1266 & 1692 & 	166 & 	480 & 906  & 817 & 	126 & 	517 & 605 & 783 & 	25 & 	205  & 857 &  1015 & 	50 & 	329 \\ 
\textbf{Dev} & {\color{white}0,}310 & 404 & 	54 & 	119  & 219 &169 & 	36 & 	141 & 148 &185 & 	11 & 	53 & 210 &252 & 	11 & 	76 \\
\textbf{Test} & {\color{white}0,} 492 & 773 & 	66 & 	155 & 328 &364 & 	63 & 	116    & 322 &317 & 	25 & 	143& 326& 407 & 	29 & 	78  \\
\bottomrule
\end{tabular}
}

\caption{Statistics of 4 datasets from ASTE-DATA-V2, where \textit{Sent.} denotes number of sentences, and \textit{Pos. Neu. Neg.} denote numbers of positive, neutral and negative triplets respectively.}
\label{tab:dataset}
\end{table*}
\begin{figure*}[t]
    \centering
    \includegraphics[width=\textwidth]{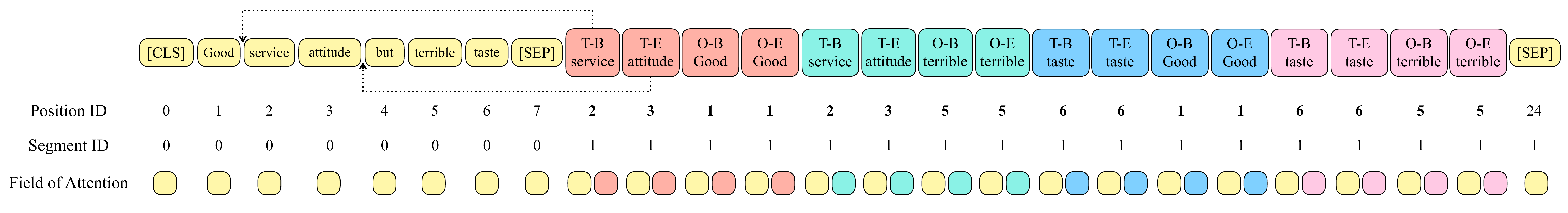}
    \caption{An example of the input structure. Note that each target/opinion span corresponds to a start tag \texttt{T-B}/\texttt{O-B} and an end tag \texttt{T-E}/\texttt{O-E}. In the last row, for each token we use square(s) to denote its attention field, meaning this token will attention to the segments with the corresponding colors.
    For example, \texttt{[CLS]} can only see the token colored yellow, while the first \texttt{T-B:service} which is colored red can see 4 more tokens that also colored red than \texttt{[CLS]}.
    }
    \label{fig:bert}
\end{figure*}
\section{Methodology}
This section will introduce our approach in detail. 
An overview of our approach is shown in Figure \ref{fig:main}.
We use a sequence labeling model to jointly extract the targets and opinions in the sentence first. Then assign perceivable pairs to all possible target-opinion pairs to mitigate the ambiguity in different target-opinion word collocation. After that, following \citet{zhong2020frustratingly}, we use the compound computation mechanism to improve computational efficiency and the restricted attention field to reduce the mutual influence.
Finally, we take the representations of perceivable pairs as the basis for matching to get the final result.
\subsection{Target and Opinion Extraction}
Our target and opinion word extraction component is a sequence labeling model based on BERT \cite{devlin2018bert} with a $\rm B, I, O, E, S$ tagging system\footnote{$\rm B, I, O, E, S$ denotes ``B-begin'', ``I-inside'', ``O-outside'', ``E-end'' and ``S-single''.} , where there are four types of labels $\rm B$, $\rm I$, $\rm E$, $\rm S$ with $\rm Target$ and $\rm Opinion$ (e.g., $\rm B-Target$), and 9 types in total for both target and opinion, as well as an unmarked label $\rm O$. 

Given an input sequence with $l$ tokens $X=w_1, w_2, ..., w_l$, we first let tokens obtain contextualized representations $R_e=\{r_{e_1}, r_{e_2}, ..., r_{e_l}\}$ through BERT. Then we do token classification to get the extraction distributions:
\begin{equation}
P^e_i = {\rm softmax}(r_{e_i}\textbf{W}_e+b_e)
\end{equation}
where $\textbf{W}_e\in \mathbb{R}^{d\times 9}$, $b_e \in \mathbb{R}^{9} $, $d$ is the dimension of encoding vectors. We extract target and opinion jointly. 

Correspondingly, we use cross-entropy as the loss function at this stage, which is defined as follows:
\begin{equation}
    \mathcal{L}_e = \sum_{i=1}^l -y^e_i{\rm log}P^e_i
\end{equation}
where 
$y^e_i$ is the one-hot vector of ground truth label.

After obtaining the classification result of each token, we will merge them into the corresponding target set $C_t$ and opinion set $C_o$ for subsequent models:
\begin{equation}
    \begin{aligned}
    C_t&=\{t_1, t_2, .. t_m\} \\
    C_o&=\{o_1, o_2, ..., o_n\}
    \end{aligned}
\end{equation}
where $m$, $n$ are the number of target and opinion words, respectively.

\subsection{Joint Matching and Sentiment Classification}

\subsubsection{Perceivable Pairs}

As discussed in Chapter \ref{chap: intro}, to well learn the word representation in a specific target-opinion pair, we need to let target/opinion words aware the target-opinion-tuple they are currently in. Therefore, we add the artificial tags named perceivable pair which correspond to the predicted targets and opinions from the first stage into the sentence. 
Indicating the span boundaries with the artificial tokens has been widely used in many Joint Entity and Relation Extraction models~\cite{zhang2019ernie,soares2019matching,zhong2020frustratingly}. We use the similar settings to enhance the connection between the specific target and opinion. 
Each perceivable pair contains four kinds of tags which correspond to the start and end of target and opinion, respectively. 
Each perceivable pair is defined as:
\begin{equation}
    A_{ij} = \{ \texttt{T-B}_{i}, \texttt{T-E}_{i}, \texttt{O-B}_{j}, \texttt{O-E}_{j}\}
\end{equation}
where $i$, $j$ represents the $i$-th target and $j$-th opinion in $X$.

\subsubsection{Compound Computations}

Commonly, more than one target or opinion may appear in a sentence. 
If we need to obtain the representations of each possible perceivable pair of targets and opinions, we will get  $m*n$  sequences. Each of them is composed of the sentence $X$ and one possible perceivable pair, as the input of the encoder.
This will bring a high computation load for the model.
Following the methods from \cite{zhong2020frustratingly}, we
use compound computation which simultaneously considers all the possible perceivable pairs in one sequence to solve this problem.

We first get all the perceivable pairs $A_{ij}$  and concatenate them together as the  segment $X_{ts}$.
\begin{equation}
\begin{aligned}
    X_{ts} = &\{ A_{11}, A_{12}, ..., A_{1n} , \\
               & ~~A_{21}, A_{22}, ..., A_{2n} , \\
               & ~~... , \\
               & ~~A_{m1}, A_{m2}, ..., A_{mn}   \}
\end{aligned}
\end{equation}
where $m,n$ are the number of targets and opinions. Then we concatenate the original sequence $X$ with $X_{ts}$ and get the new sequence:
\begin{equation}
    X_p= X + X_{ts} 
\end{equation}

At the same time, the perceivable pairs keep their corresponding position information by sharing their position embeddings with the boundary tokens of the corresponding span:
\begin{equation}
\begin{aligned}
    {\rm Position}(\texttt{T-B}_i)&= {\rm Position}(t_{i-{\rm start}}) \\
    {\rm Position}(\texttt{T-E}_i)&= {\rm Position}(t_{i-{\rm end}}) \\
    {\rm Position}(\texttt{O-B}_j)&= {\rm Position}(o_{j-{\rm start}}) \\
    {\rm Position}(\texttt{O-E}_j)&= {\rm Position}(o_{j-{\rm end}})
\end{aligned}
\end{equation}

Finally, different from~\cite{zhong2020frustratingly},
we use different segment id for each token in the sequence $X_p$ to distinguish the original sentence and newly added perceivable pairs, as shown in Figure \ref{fig:bert}.
\subsubsection{Restricted Attention Field}
Since multiple perceivable pairs occur simultaneously in the same sequence,
they interfere each other and may confuse the model. To reduce their interference, we also adopt the restricted attention field from~\cite{zhong2020frustratingly} to let different types of tokens have different activated attention fields.

For each token in $X_{ts}$, its attention field includes its corresponding perceivable pair and the original sequence $X$.
Meanwhile, each token in the segment $X_{ts}$ is not visible to tokens in $X$.  
Figure \ref{fig:bert}
shows an example sequence composed of the original sentence and the four perceivable pairs,  which are distinguished by five different colors.
In the last row, for each token we use  square(s) to denote its attention field, meaning this token will attend to the segments with the corresponding colors. 
Formally, the attention field of tokens can be defined as follows:
\begin{equation}
    {\rm AttnField}(w_i)=\left\{
    \begin{aligned}
        & X, & w_i \notin X_{ts}\\
        & X \cup A_{ij}, & w_i \in X_{ts}
    \end{aligned}
    \right.
\end{equation}
where $w_i, w_j \in A_{ij}$ .
\subsubsection{Matching} 
We put the modified sequence $X_{p}$ into BERT 
to obtain the correlation-enhanced representation:
\begin{equation}
    \begin{aligned}
        R_{p}  = \{& h_\texttt{[CLS]}, h_1, ..., \\ &h_\texttt{[SEP]}, ..., H_{ij}, ..., h_\texttt{[SEP]} \}
    \end{aligned}
\end{equation}
where $H_{ij}  = \{ h_{\texttt{T-B}_{i}}, h_{\texttt{T-E}_{i}}, h_{\texttt{O-B}_{j}}, h_{\texttt{O-E}_{j}} \}$.

After that, we fuse the representation of the target $i$ and opinion $j$ in $A_{ij}$ as the representation of the perceivable pair:
\begin{equation}
    r_{ij} = [h_{\texttt{T-B}_i}; h_{\texttt{O-B}_j}]
\end{equation}
where $;$ represents vector concatenation.

Finally, the representation $r_{ij}$ is used to predict their matching result and sentiment polarity:
\begin{equation}
    P^m_{ij} = {\rm softmax}(r_{ij}\textbf{W}_m+b_m)
\end{equation}
where $\textbf{W}_m \in \mathbb{R}^{2d\times 4}$, $b_m \in \mathbb{R}^{4}$. Each pair of target and opinion will be classified into four categories:
\begin{equation}
    L_m=\{\texttt{POS},\texttt{NEU}, \texttt{NEG}\} \cup \{\texttt{O}\}
\end{equation}
where the label \texttt{O} indicates that this pair of words does not match.
Here, we use cross-entropy as the loss function. The total loss of a sentence is the sum of each possible target-opinion pair's loss in the sentence, which is defined as follows:
\begin{equation}
    \mathcal{L}_m = -\sum_{i=1}^p \sum_{j=1}^q y^m_{ij}{\rm log}P^m_{ij}
\end{equation}
where $p$ and $q$ are the number of targets and opinions in the sentence, respectively. $y^m_{ij}$ is the one-hot vector of the ground-truth label of matching and sentiment polarity.

\begin{table*}[t]
    \centering
    \scalebox{0.77}{
    \begin{tabular}{lccc|ccc|ccc|ccc}
    \toprule
      \multirow{2}{*}{\textbf{Models}} & \multicolumn{3}{c}{\texttt{14Rest}} & \multicolumn{3}{c}{\texttt{14Lap}} & \multicolumn{3}{c}{\texttt{15Rest}}  & \multicolumn{3}{c}{\texttt{16Rest}}    \\ 
      & $P.$ & $R.$ & $F_1$& $P.$ & $R.$ & $F_1$& $P.$ & $R.$ & $F_1$ & $P.$ & $R.$ & $F_1$ \\ \midrule
       
        CMLA+ & 39.18 & 47.13 & 42.79 & 30.09 & 36.92 & 33.16 &  34.56 & 39.84 & 37.01 & 41.34 & 42.10 & 41.72          \\
        RINANTE+ &  31.42 & 39.38 & 34.95 &  21.71 & 18.66 & 20.07  &  29.88 & 30.06 & 29.97 &  25.68 & 22.30 & 23.87  \\
        Li-unified-R &  41.04 & 67.35 & 51.00 & 40.56 & 44.28 & 42.34 & 44.72 & 51.39 & 47.82 &  37.33 & 54.51 & 44.31 \\
        \newcite{peng2020knowing} & 43.24 & 63.66 & 51.46 & 37.38 & 50.38 & 42.87 & 48.07 & 57.51 & 52.32 & 46.96 & 64.24 & 54.21 \\ 
        OTE-MTL & 63.07 & 58.25	& 60.56 & 54.26 & 41.07 & 46.75 &  60.88 & 42.68 & 50.18 &  65.65 & 54.28	&59.42  \\
        GTS-BiLSTM & 71.41 & 53.00 & 60.84 & 58.02 & 40.11 & 47.43 & 64.57 & 44.33 & 52.57 & 70.17 & 55.95 & 62.26 \\
        JET$^t$ & 66.76&49.09& {56.58}&  52.00& 35.91 &42.48 & 59.77&42.27& {49.52}& 63.59&50.97&{56.59}\\
        {JET}$^o$ & 61.50 &55.13&{58.14}&53.03	&33.89&	41.35 &64.37&	44.33&	{52.50}&  70.94 & 57.00 & {63.21}\\
        \midrule
        GTS$_{\scriptscriptstyle \text{+ BERT}}$ & 67.25 & 69.22 & \textbf{68.22} & 58.54 & 50.65 & 54.30 & 60.69 & 60.54 & \textbf{60.61} & 67.39 & 66.73 & 67.06 \\
        JET$^t_{\scriptscriptstyle \text{+ BERT}}$ &
         63.44 & 54.12 & 58.41 &  53.53 & 43.28 & 47.86 &  {68.20} & 42.89 & 52.66 & 65.28 & 51.95 & 57.85 \\
        JET$^o_{\scriptscriptstyle \text{+ BERT}}$ &
        {70.56} & 55.94 & 62.40 & 55.39 & 47.33 & 51.04 & 64.45 & 51.96 & 57.53 & {70.42} & 58.37 & 63.83 \\
        \midrule 
        \textbf{Ours} & 63.59&73.44&{68.16}&57.84&59.33&\textbf{58.58}&54.53&63.30&{58.59}&63.57&71.98&\textbf{67.52}\\
  \bottomrule
    \end{tabular}
   } 
    \caption{Main results. $P.$ and $R.$ are \textbf{P}recision and \textbf{R}ecall respectively. The best $F_1$ for each dataset have been \textbf{bolded}.}
    \label{tab:main_results}
\end{table*}
\section{Experiments and Analysis}
\subsection{Datasets}
We utilize datasets created by \cite{xu2020position} named ASTE-DATA-V2 in our experiment. Compared to ASTE-Data-V1 proposed by \cite{peng2020knowing}, the V2 version contains cases where one target/opinion is associated with multiple opinions/targets, which is very common in actual scenarios.
The overview of datasets is listed in Table \ref{tab:dataset}. The division of the dataset is consistent with that of \cite{xu2020position}.
\subsection{Baselines}
We compare our method with the following baseline models, some pipeline-based methods are modified by \cite{peng2020knowing} for this task:
\begin{itemize}

     \item \textbf{CMLA+} is modified from CMLA \cite{wang2017coupled}. 
     CMLA uses the attention mechanism to capture the relationship between words and jointly extract target and opinion, and CMLA+ further adds an MLP on CMLA to determine whether a triplet is correct in the matching stage.
     
     \item \textbf{RINANTE+} is modified from RINANTE \cite{dai2019neural}. RINANTE is based on LSTM-CRF and fuses rules as weak supervision to capture words' dependency relations in a sentence. The way RINANTE+ determines the correctness of a triplet is the same as CMLA+.
     
     \item \textbf{Li-unified-R}  \cite{li2019unified} extracts targets, sentiment and opinion spans respectively based on a multi-layer LSTM neural architecture. 
     The way Li-unified-R determines the correctness of a triplet is the same as CMLA+.
     
     \item \newcite{peng2020knowing} co-extracts targets with sentiment, and opinion spans like \cite{li2019unified}, and uses GCN to capture dependency information to enhance the co-extraction. The way to determine the correctness of a triplet is also the same as CMLA+.
     
     \item \textbf{OTE-MTL} \cite{zhang2020multi} is a multi-task learning framework to extract aspect terms and opinion terms jointly and simultaneously parses sentiment dependencies between them. 
     
     \item \textbf{GTS} \cite{wu-etal-2020-grid} address the ASTE task in an end-to-end fashion with one unified grid tagging task. They designed an gird inference strategy to exploit mutual indication for more accurate extractions. Their models have two variants that use Glove and BERT to initialize the encoder layer.
     
     \item \textbf{JET} \cite{xu2020position} converts the ASTE task into several sequence labeling subtasks. 
     In this method, JET$^t$ takes the target words as the labeling object, and the label includes the span of the target, the sentiment polarity, and the offset of the paired opinion. JET$^o$ is similar to JET$^t$ expect the labeling objectives are opinion words.
     
\end{itemize}
\begin{table*}[t]
    \centering
    \scalebox{0.73}{
    \begin{tabular}{lccc|ccc|ccc|ccc}
    \toprule
      \multirow{2}{*}{\textbf{Models}} & \multicolumn{3}{c}{\texttt{14Rest}} & \multicolumn{3}{c}{\texttt{14Lap}} & \multicolumn{3}{c}{\texttt{15Rest}}  & \multicolumn{3}{c}{\texttt{16Rest}}    \\ 
      & $P.$ & $R.$ & $F_1$& $P.$ & $R.$ & $F_1$& $P.$ & $R.$ & $F_1$ & $P.$ & $R.$ & $F_1$ \\ \midrule
        \textbf{Original Framework} &   63.59&73.44&68.16&57.84&59.33&58.58&54.53&63.30&58.59&62.90&72.57&67.39 \\
        \midrule
        \textbf{a) rm. Start Tags} & 57.95&70.42&63.58&48.59&57.30&52.59&50.00&59.59&54.37&57.82&71.21&63.82 \\
        \textbf{b) rm. End Tags} & 55.78&68.41&61.46&43.57&55.08&48.65&46.19&60.00&52.20&58.53&71.40&64.33 \\
        \textbf{c) rm. All Tags} & 40.09&64.29&49.38&41.62&56.01&47.75&36.33&55.05&43.77&42.72&68.48&52.62 \\
        \textbf{d) rm. Tag Segment} & 51.32&72.33&60.04&43.24&59.70&50.16&47.68&61.44&53.69&55.11&71.40&62.20 \\
        \textbf{e) rm. Tag Restricted Attn.} & 46.25&73.74&56.84&40.98&61.74&49.26&36.73&62.47&46.26&47.00&73.15&57.23 \\
        \textbf{f) rm. All Restricted Attn.} & 43.51&73.24&54.59&41.74&59.33&49.01&34.43&60.21&43.81&43.28&71.40&53.89\\
  \bottomrule
    \end{tabular}
   } 
    \caption{Results of ablation study, and \textit{rm.} in this table means remove. The settings of all experiments are consistent with the main experiment. }
    \label{tab:analysis_results}
\end{table*}
\subsection{Evaluation Metrics}
We use precision, recall, and micro $F_1$ as the evaluation metrics of triple extraction, which is consistent with the previous works. Only if all the elements of a triplet, i.e., target, opinion, and their corresponding sentiment polarity are correct, it will be regarded as a correct result in evaluation.

It is noted that the results of some models are directly taken from~\cite{xu2020position}.
And in the original paper of OTE-MTL \cite{zhang2020multi} and GTS  \cite{wu-etal-2020-grid}, they use ASTE-DATA-V1 datasets for training and evaluation. So we re-run their models on ASTE-DATA-V2 datasets. All the baselines implemented by us use their default hyper-parameters and report the best results of 5 different random seeds for a fair comparison. 
\subsection{Implementation Details}
We implement our models based on HuggingFace’s \texttt{Transformers} library \cite{wolf-etal-2020-transformers} and use \texttt{bert-base-uncased} \cite{devlin2018bert} as the base encoders. We optimize our models with a learning rate of 5e-5 by Adam, random seed range of $[1, 5]$. The max sequence length is set to 256, and the batch size is 8. We train 3 epochs for target and opinion extraction stage and 10 epochs for the matching stage for all the experiments.
We select the final model based on the performance on development set with the hyperparameters above, and report its results on the test set.
All experiments are conducted on a Linux server with Intel Xeon E5-2680, 256G of RAM and Nvidia RTX 3090.

\subsection{Method Comparison}
Table \ref{tab:main_results} reports the experimental results of our model against other baseline methods. Our model achieves state-of-the-art results on \texttt{14Lap}, \texttt{16Rest}, close to \textbf{GTS} in \texttt{14Rest}, and slightly behind \textbf{GTS} in \texttt{15Rest}. And our model achieves the best overall $F_1$ of the four datasets, which prove the effectiveness of the model.

First, due to the strong expressive ability of BERT,
the performance of the model can be improved. 
More importantly, our model's performance is much better than other models with BERT like  {JET}$_{\scriptscriptstyle \text{+ BERT}}$. 
Because our model can traverse all possible target-opinion pairs to match, which overcome the shortcomings of tagging-based models that cannot handle one-to-many situations.

Then, compared with OTE-MTL, which is also a matching-based method, the performance of our method is significantly better. OTE-MTL only selects the terms' original representation as the basis for matching, which cannot reflect target-opinion pairs' correlation. This may be the actual reason for the performance difference. 

Finally, GTS combines the representations of target and opinion and generates specific representations for each target-opinion pair like ours. But the word representations used for combining still remain the same for different target-opinion pairs. While our model generates specific word representations for the same word in each possible pair, then combines those specific word representations to generate pair representations to establish correlations within target-opinion pairs, which results in performance improvement.
\subsection{Ablation Study}
To verify the effects of the components in our framework, we conduct some ablation experiments. The results are shown in Table \ref{tab:analysis_results}. 

We first verified the effectiveness of perceivable pairs by removing the start tag and the end tag of perceivable pairs, respectively. The experimental results are given in Table \ref{tab:analysis_results} a) and b). It can be seen that the performance has shown some decline. This shows that perceivable pairs are fully effective in indicating complete term spans. 

After that, we completely remove the perceivable pairs from the sequence and only use the representation of the term for matching like previous matching-based methods. The results are shown in Table \ref{tab:analysis_results} c). We notice that the performance has dropped more significantly than the previous two experiments, which shows the perceivable pairs inserted into the sequence can establish the correlation between them by making target/opinion words aware the target-opinion-tuple they are current in.

Finally, we removed the tag segment's special segment id and merged it with the original sequence into one segment. The experimental results are shown in d). Since perceivable pairs are artificially added tokens, they may confuse the model if they are not distinguished from the original input.

Then we conducted two experiments on the restricted attention field. We first removed the restricted attention field in the tag segment. All perceivable pairs in tag segment can be seen by other pairs, but the tokens in the original sequence remain the same as before.
The experimental results are shown in Table \ref{tab:analysis_results} e). Performance has dropped significantly. 
As discussed before, perceivable pairs are effective because they establish a strong correlation between target-opinion pairs. After removing the attention field restriction in the tag segment, tags in perceivable pairs can see all other tags in the sentence instead of being limited to the specific target-opinion pair. Thus the correlation is significantly weakened. 
Therefore, the performance of the model is significantly declined.

Then we go one step further and remove all restricted attention field, so all tokens in the sequence could see other tokens without any restriction. The experimental results are shown in f), exhibiting a more obvious decline than e). 
At this time, the pairwise correlation of the target-opinion pair no longer exists, and the original tokens see a large number of artificial tokens, which also affect their representation.

\subsection{Case study on Perceivable Pairs}
\begin{figure}[t]
    \centering
    \includegraphics[width=0.45\textwidth]{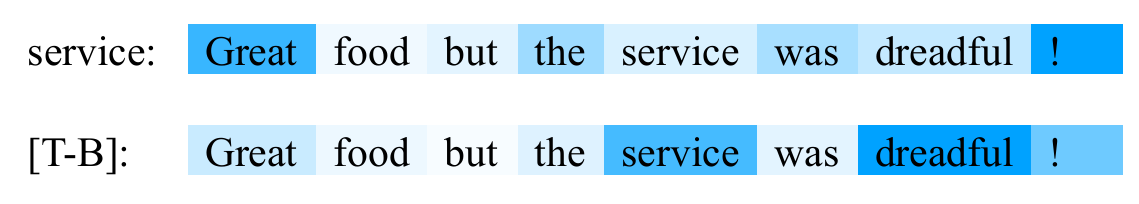}
    \caption{Visual comparison of the attention weight between target word (\textit{service}) and corresponding start tag in perceivable pairs. }
    \label{fig:case_study}
\end{figure}
\begin{figure}[t]
    \centering
    \includegraphics[width=0.45\textwidth]{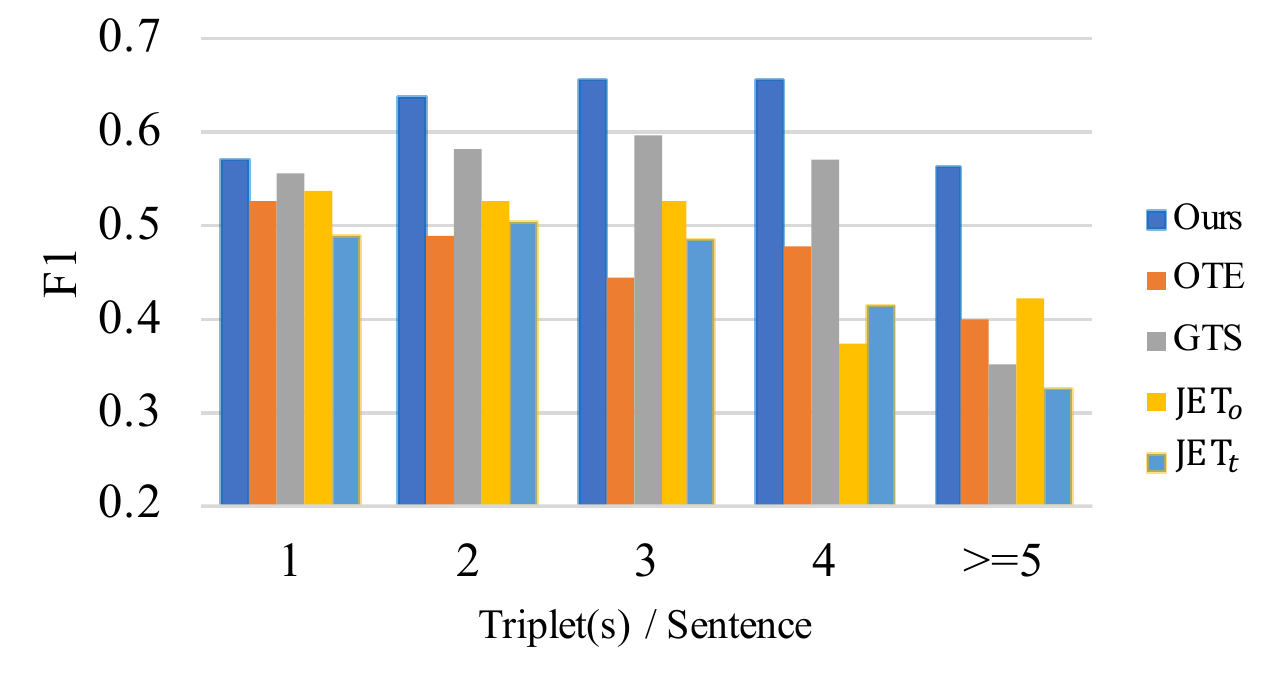}
    \caption{Performance comparison in complex situations with multiple triplets in a sentence. Due to limited space, we only show the results of the \texttt{14Lap} dataset. All the models mentioned in this figure use BERT as the encoder.}
    \label{fig:data_complex}
\end{figure}

\begin{table}[t]
    \centering
    
    \scalebox{0.7}{
    \begin{tabular}{lccc|ccc}
    
    \toprule
      \multirow{2}{*}{\textbf{Models}}  & \multicolumn{3}{c}{\texttt{14Lap}} & \multicolumn{3}{c}{\texttt{15Rest}} \\
      & $P.$ & $R.$ & $F_1$& $P.$ & $R.$ & $F_1$  \\ \midrule
      Ours &70.18&59.07&64.15&68.36&69.54&68.95 \\
      \midrule
      OTE-MTL$_{\scriptscriptstyle \text{+ BERT}}$ & 64.33 & 42.15 & 50.93 &59.37 &43.68& 50.33 \\
      JET$^o_{\scriptscriptstyle \text{+ BERT}}$& 60.49&37.55&46.34&67.59&41.95&51.77\\
      JET$^t_{\scriptscriptstyle \text{+ BERT}}$ & 59.62&35.63&44.60&68.00&29.31&40.96\\
  
  \bottomrule
    \end{tabular}
   } 
    
    \caption{Performance comparison in complex situations with one-to-many correspondence. All the models mentioned in this table use BERT as the encoder. Due to limited space, we only list the results of two datasets. Similar results can be seen on other datasets.}
    \label{tab:ont_to_many}
    
\end{table}
To better explore the effect of perceivable pairs in the model, we conducted a case study. We feed the sequence with perceivable pairs into BERT, and check the attention weight of the target word (\textit{service}) and its corresponding start tag in perceivable pair respectively. 
The result is shown in Figure \ref{fig:case_study}. It turns out that the former will pay attention to all the opinion words (\textit{Great} and \textit{dreadful}) in the sentence, but only \textit{dreadful} really needs to be focused. While the latter using perceivable pairs focus on the truly important words (\textit{service} and \textit{dreadful}). 
Moreover, it can be seen from the attention weight distribution that those tags in perceivable pair establish a strong correlation between the target-opinion pair.
The final sentiment classification result using only the original 
representation of the \textit{service} in the sentence is incorrectly classified as positive. In contrast, the classification based on perceivable pair is correct and marked as negative. 
The comparison of the two results also shows the actual effect of the perceivable pairs.
\subsection{Model performance in complex situations}
We examined the performance comparison between our method and other baseline models in complex situations where there are multiple triplets in a sentence. The results are shown in Figure \ref{fig:data_complex}. As the number of triplets increases, the performance gap between our framework and other models becomes more obvious. Since our method uses different perceivable pairs to model all possible target-opinion pairs, as the number of triplets in the sentence increases, our method will not be greatly affected.

We also investigated the one-to-many problem in which a word appears in multiple triplets in a sentence. We extract this part of the data for experiments, and the results are shown in Table \ref{tab:ont_to_many}. It can be seen that in the one-to-many situation, our model also has significant performance advantages. Since our model utilizes the restricted attention field to isolate different target-opinion combinations of the same word, it will not be affected much in the complex scenario like one-to-many.
\section{Conclusion}
This paper utilize a two-stage framework for ASTE tasks. We first extract the target and opinion words by sequence labeling. 
Then, we use the perceivable pairs at the second stage to make target/opinion words aware the target-opnion-tuples they are currently in, which enhance the target-opinion correlation and further improve matching accuracy.
Moreover, we use the compound computations to accelerate training and inference, and restricted attention field to reduce mutual interference. Results from detailed experiments show that our method achieves good performance on four datasets and gets more significant performance advantages in complex situations.

\bibliography{anthology,custom}
\bibliographystyle{acl_natbib}

\end{document}